# LQ-Nets: Learned Quantization for Highly Accurate and Compact Deep Neural Networks

Dongqing Zhang*, Jiaolong Yang*, Dongqiangzi Ye*, and Gang Hua

Microsoft Research
zdqzeros@gmail.com jiaoyan@microsoft.com eowinye@gmail.com ganghua@microsoft.com

**Abstract.** Although weight and activation quantization is an effective approach for Deep Neural Network (DNN) compression and has a lot of potentials to increase inference speed leveraging bit-operations, there is still a noticeable gap in terms of prediction accuracy between the quantized model and the full-precision model. To address this gap, we propose to jointly train a quantized, bit-operation-compatible DNN and its associated quantizers, as opposed to using fixed, handcrafted quantization schemes such as uniform or logarithmic quantization. Our method for learning the quantizers applies to both network weights and activations with arbitrary-bit precision, and our quantizers are easy to train. The comprehensive experiments on CIFAR-10 and ImageNet datasets show that our method works consistently well for various network structures such as AlexNet, VGG-Net, GoogLeNet, ResNet, and DenseNet, surpassing previous quantization methods in terms of accuracy by an appreciable margin. Code available at https://github.com/Microsoft/LQ-Nets

**Keywords:** Deep Neural Networks · Quantization · Compression

## 1 Introduction

Deep neural networks, especially the deep convolutional neural networks, have achieved tremendous success in computer vision and the broader artificial intelligence field. However, the large model size and high computation cost remain great hurdles for many applications, especially on some constrained devices with limited memory and computational resources.

To address this issue, there has been a surge of interests recently in reducing the model complexity of DNNs. Representative techniques include quantization [6, 34, 29, 54, 52, 21, 39, 53, 22, 3, 55, 9, 18], pruning [13, 12, 17, 36], low-rank decomposition [7, 8, 24, 27, 38, 51, 49], hashing [4], and deliberate architecture design [23, 19, 50]. Among these approaches, quantization based methods represent the network weights with very low precision, thus yielding highly compact DNN models compared to their floating-point counterparts. Moreover, it has been shown that if both the network weights and activations are properly quantized, the convolution operations can be efficiently computed via bitwise operations [39, 21], enabling fast inference without GPU.

---

* Authors contributed equally. This work was done when DY was an intern at MSR.

Notwithstanding the promising results achieved by the existing quantization-based methods [6, 34, 29, 54, 52, 21, 39, 53, 22, 3, 55, 9, 18], there is still a sizeable accuracy gap between the quantized DNNs and their full-precision counterparts, especially when quantized with extremely low bit-widths such as 1 bit or 2 bits. For example, using the state-of-the-art method of [3], a 50-layer ResNet model [15] with 1-bit weights and 2-bit activations can achieve 64.6% top-1 image classification accuracy on ImageNet validation set [40]. However, the full-precision reference is 75.3% [15], i.e., the absolute accuracy drop induced by quantization is as large as 10.7%.

This work is devoted to pushing the limit of network quantization algorithms to achieve better accuracy with low precision weights and activations. We found that existing methods often use simple, hand-crafted quantizers (e.g., uniform or logarithmic quantization) [11, 31, 53, 22, 37, 52] or otherwise pre-computed quantizers fixed during network training [3]. However, one can never be sure that the simple quantizers are the best choices for network quantization. Moreover, the distributions of weights and activations in different networks and even different network layers may differ a lot. We believe a better quantizer should be made adaptive to the weights and activations to gain more flexibility.

To this end, we propose to jointly train a quantized DNN and its associated quantizers. The proposed method not only makes the quantizers learnable, but also renders them compatible with bitwise operations so as to keep the fast inference merit of properly-quantized neural networks. Our quantizer can be optimized via backpropagation in a standard network training pipeline, and we further propose an algorithm based on quantization error minimization which yields better performance. The proposed quantization can be applied to both network weights and activations, and arbitrary bit-width can be achieved. Moreover, layer-wise quantizers with unshared parameters can be applied to gain further flexibility. We call the networks quantized by our method the "LQ-Nets".

We evaluate our LQ-Nets with image classification tasks on the CIFAR-10 [25] and ImageNet [40] datasets. The experimental results show that they perform remarkably well across various network structures such as AlexNet [26], VGG-Net [41], GoogLeNet [42], ResNet [15] and DenseNet [20], surpassing previous quantization methods by a wide margin.

## 2   Related Work

A large number of works have been devoted to reducing DNN model size and improving inference efficiency for practical applications. We briefly review the existing approaches as follows.

**Compact network design:** To achieve fast inference, one strategy is to carefully design a compact network architecture [32, 42, 23, 19, 50]. For example, Network in Network [32] enhanced the local modeling via the micro networks and replaced the costly fully-connected layer by global average pooling. GoogLeNet [42] and SqueezeNet [23] utilized 1×1 convolution layers to compute reductions before the expensive 3×3 or 5×5 convolutions. Similarly, ResNet [15] applied "bottleneck" structures with 1×1 convolutions when training deeper nets with enormous parameters. The recently proposed computation-efficient network structures MobileNet [19] and ShuffleNet [50] employed depth-wise convolution or group convolution advocated in [5, 48] to reduce the computation cost.

**Network parameter reduction:** Considerable efforts have been devoted to reducing the number of parameters in an existing network [13, 12, 17, 36, 7, 8, 24, 27, 38, 51, 49, 4, 35, 28, 45, 10, 46]. For example, by exploiting the redundancy of the filters weights, some methods substitute the pre-trained weights using their low-rank approximations [7, 8, 24, 27, 38, 51, 49]. Connection pruning was investigated in [13, 12] to reduce the parameters of AlexNet and VGG-Net, where significant reduction was achieved on fully-connected layers. Promising results on modern network architectures such as ResNet were achieved recently by [17, 36]. Another similar technique is to regularize the network by structured sparsity to obtain a hardware-friendly DNN model [35, 28, 45]. Some other approaches such as hashing and vector quantization [44] have also been explored to reduce DNN model complexity [4, 10, 46].

**Network quantization:** Another category of existing methods, which our method also belongs to, train low-precision DNNs via quantization. These methods can be further divided into two subcategories: those performing quantization on *weights only* versus *both weights and activations*.

For weight-only quantization methods, Courbariaux et al. [6] constrained the weights to only two possible values of −1 and 1 (i.e., binarization or one-bit quantization). They obtained promising results on small datasets using stochastic binarization. Rastegari et al. [39] later demonstrated that deterministic binarization with optimized scale factors to approximate the full-precision weights work better on deeper network structures and larger datasets. To obtain better accuracy, ternary and other multi-bit quantization schemes were explored in [34, 29, 54, 52, 9, 18]. It was shown in [52] that quantizing a network with five bits can achieve similar accuracy to its 32-bit floating-point counterpart by incremental group-wise quantization and re-training.

In the latter regard, Hubara et al. [21] and Rastegari et al. [39] proposed to binarize both weights and activations to −1 and +1. This way, the convolution operations can be implemented by efficient bit-wise operations for substantial speed-up. To address the significant accuracy drop, multi-bit quantization was further studied in [53, 22, 37, 43, 30, 33]. A popular choice of the quantization function is the uniform quantization [53, 22]. Miyashita et al. [37] used logarithmic quantization and improve the inference efficiency via the bitshift operation. Cai et al. [3] proposed to binarize the network weights while quantize the activations using multiple bits. A single activation quantizer computed by fitting the probability density function of a half-wave Gaussian distribution is applied to all network layers and fixed during training. In the multi-bit quantization methods of Tang et al. [43] and Li et al. [30], each bit is used to binarize the residue approximation error from previous bits.

Our proposed method can quantize both the weights and the activations with arbitrary bit-widths. Different from most of the previous methods, our quantizer is adaptively learned during network training.

## 3    LQ-Nets: Networks with Learned Quantization

In this section, we first briefly introduce the goal of neural network quantization. Then we present the details of our quantization method and how to train a quantized DNN model with it in a standard network training pipeline.

### 3.1    Preliminaries: Network Quantization

The main operations in deep neural networks are interleaved linear and non-linear transformations, expressed as

$$z = \sigma(\mathbf{w}^{\mathrm{T}}\mathbf{a}), \tag{1}$$

where $\mathbf{w} \in \mathbb{R}^N$ is the weight vector, $\mathbf{a} \in \mathbb{R}^N$ is the input activation vector computed by the previous network layer, $\sigma(\,\cdot\,)$ is a non-linear function, and $z$ is the output activation.[1] The convolutional layers are composed by multiple convolution filters $\mathbf{w}_i \in \mathbb{R}^{C \cdot H \cdot W}$, where $C$, $H$ and $W$ are the number of convolution filter channels, kernel height, and kernel width, respectively. Fully-connected layers can be viewed as a special type of convolutional layer. Modern deep neural networks often have millions of weight parameters, which incur large memory footprints. Meanwhile, the large numbers of inner product operations between the weights and feature vectors lead to high computation cost. The memory and computation costs are great hurdles for many applications on resource-constrained devices such as mobile phones.

The goal of network quantization is to represent the floating-point weights $\mathbf{w}$ and/or activations $\mathbf{a}$ with few bits. In general, a quantization function is a piecewise-constant function which can be written as

$$Q(x) = q_l, \text{ if } x \in (t_l, t_{l+1}], \tag{2}$$

where $q_l$, $l = 1, ..., L$ are the quantization levels and $(t_l, t_{l+1}]$ are quantization intervals. The quantization function maps all the input values within a quantization interval to the corresponding quantization level, and a quantized value can be encoded by only $\log_2 L$ bits. Perhaps the simplest quantizer is the sign function used for binary quantization [21, 39]: $Q(x) = +1$ if $x \geq 0$ or $-1$ otherwise. For quantization with 2 or more bits, the most commonly used quantizer is the uniform quantization function where all the quantization steps $q_{l+1} - q_l$ are equal [53, 22]. Some methods use logarithmic quantization which uniformly quantizes the data in the *log*-domain [37].

Quantizing the network weights can generate highly compact and memory-efficient DNN models: using $n$-bit encoding, the compression rate is $\frac{32}{n}$ or $\frac{64}{n}$ compared to the 32-bit or 64-bit floating point representation. Moreover, if both weights and activations are quantized properly, the inner product in Eq. (1) can be computed by bitwise operations such as *xnor* and *popcnt*, where *xnor* is the exclusive-not-or logical operation and *popcnt* counts the number of 1's in a bit string. Both the two operations can process at least 64 bits in one or few clock cycle on most general computing platforms such as CPU and GPU, which potentially leads to 64× speedup.[2]

---

[1] For brevity, we omit the bias term in Eq. (1).

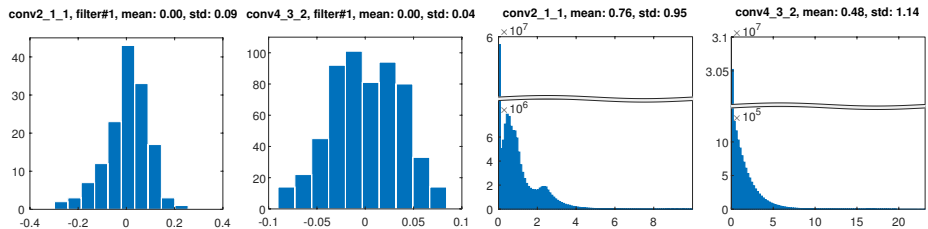

**Fig. 1:** Distributions of weights (left two columns) and activations (right two columns) at different layers of the ResNet-20 network trained on CIFAR-10. All the test-set images are used to get the activation statistics.

### 3.2 Learnable Quantizers

An optimal quantizer should yield minimal quantization error for the input data distribution:

$$Q^*(x) = \arg\min_Q \int p(x)(Q(x) - x)^2 dx, \tag{3}$$

where $p(x)$ is the probability density function of $x$. We can never be sure if the popular quantizers such as a uniform quantizer are the optimal selections for the network weights and activations. In Fig. 1 we present the statistical distributions of the weights and activations (after batch normalization (BN) and Rectified Linear Unit (ReLU) layers) in a trained floating-point network. It can be seen that the distributions can be complex and differ across layers, and a uniform quantizer is not optimal for them. Of course, if we train a quantized network the weight and activation distributions may change. But again we can never be sure if any pre-defined quantizer is optimal for our task, and an improper quantizer can easily jeopardize the final accuracy.

To get better network quantizers and improve the accuracy of a quantized network, we propose to jointly train the network and its quantizers. The insight behind is that if the optimizers are learnable and optimized through network training, they can not only minimize the quantization error, but also adapt to the training goal thus improving the final accuracy. A naive way to train the quantizers would be directly optimizing the quantization levels $\{q_l\}$ in network training. However, such a naive strategy would render the quantization functions not compatible with bitwise operations, which is undesired as we want to keep the fast inference merit of quantized neural networks.

---

[2] We refer the readers to [21, 39] for more details regarding the bitwise operations and speed-up analysis on different hardware platforms.

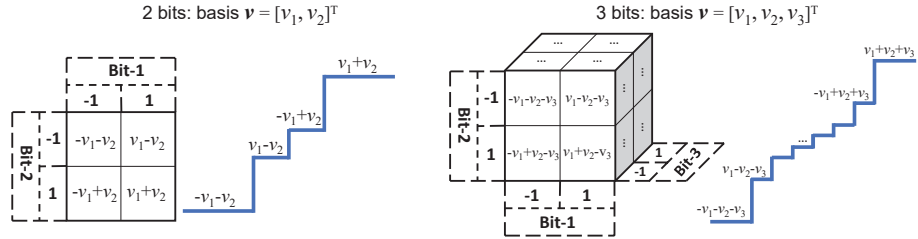

**Fig. 2:** Illustration of our learnable quantizer on the 2-bit (left) and 3-bit (right) cases. For each case, the left figure shows how quantization levels are generated by the basis vector, and the right figure illustrates the corresponding quantization function.

To resolve this issue, we need to confine our quantization functions into a *subspace* which is compatible with bitwise operations. But how to confine the quantizers into such a space during training? Our solution is inspired by the uniform quantization which is bit-op compatible (see [53]). The uniform quantization essentially maps floating-point numbers to their nearest fixed-point integers with a normalization factor, and the key property for it to be bit-op-compatible is that the quantized values can be decomposed by a linear combination of the bits. Specifically, an integer $q$ represented by a $K$-bit binary encoding is actually the inner product between a basis vector and the binary coding vector $\mathbf{b} = [b_1, b_2, ..., b_K]^\mathrm{T}$ where $b_i \in \{0, 1\}$, i.e.,

$$q = \left\langle \begin{bmatrix} 1 \\ 2 \\ ... \\ 2^{K-1} \end{bmatrix}, \begin{bmatrix} b_1 \\ b_2 \\ ... \\ b_K \end{bmatrix} \right\rangle. \tag{4}$$

In order to learn the quantizers while keeping them compatible with bitwise operations, we can simply learn the basis vector which consists of $K$ scalars.

Concretely, our learnable quantization function is simply in the form of

$$Q_{\mathrm{ours}}(x, \mathbf{v}) = \mathbf{v}^\mathrm{T} \mathbf{e}_l, \qquad \text{if } x \in (t_l, t_{l+1}], \tag{5}$$

where $\mathbf{v} \in \mathbb{R}^K$ is the learnable floating-point basis and $\mathbf{e}_l \in \{-1, 1\}^K$ for $l = 1, \ldots, 2^K$ enumerates all the $K$-bit binary encodings from $[-1, \ldots, -1]$ to $[1, \ldots, 1]$.[3] For a $K$-bit quantization, the $2^K$ quantization levels are generated by $q_l = \mathbf{v}^\mathrm{T} \mathbf{e}_l$ for $l = 1, \ldots, 2^K$. Given $\{q_l\}$ and assuming $q_1 < q_2 < ... < q_{2^K}$, it can be easily derived that for any $x$, the optimal $\{t_l\}$ minimizing the error in Eq. (3) are simply $t_l = (q_{l-1} + q_l)/2$ for $l = 2, ..., 2^K$ (note $t_1 = -\infty$ and $t_{2^K+1} = +\infty$). Figure 2 illustrates our quantizer with the 2-bit and 3-bit cases.

---

[3] Note that $\mathbf{e}_i$ can be either $\{0, 1\}$ encodings or $\{-1, 1\}$ encodings, both of which can yield quantizers compatible with bitwise operations. In our implementation, we adopt the $\{-1, 1\}$ encoding for weights and $\{0, 1\}$ encoding for activations. For convenience we will use the $\{-1, 1\}$ encoding in the remaining text as the example.

We now show how the inner products between our quantized weights and activations can be computed by bitwise operations. Let a weight vector $\mathbf{w} \in \mathbb{R}^N$ be encoded by the vectors $\mathbf{b}_i^w \in \{-1, 1\}^N$, $i = 1, \ldots, K_w$ where $K_w$ is the bit-width for weights and $\mathbf{b}_i^w$ consists of the encoding of the $i$-th bit for all the values in $\mathbf{w}$. Similarly, for activation vector $\mathbf{a} \in \mathbb{R}^N$ we have $\mathbf{b}_j^a \in \{-1, 1\}^N$, $j = 1, \ldots, K_a$. It can be readily derived that

$$Q_{\text{ours}}(\mathbf{w}, \mathbf{v}^w)^{\mathrm{T}} Q_{\text{ours}}(\mathbf{a}, \mathbf{v}^a) = \sum_{i=1}^{K_w} \sum_{j=1}^{K_a} v_i^w v_j^a (\mathbf{b}_i^w \odot \mathbf{b}_j^a) \qquad (6)$$

where $\mathbf{v}^w \in \mathbb{R}^{K_w}$ and $\mathbf{v}^a \in \mathbb{R}^{K_a}$ are the learned basis vectors for the weight and activation quantizers respectively, and $\odot$ denotes the inner product with bitwise operations *xnor* and *popcnt*.

In practice, we apply layer-wise quantizers for activations (i.e., one quantizer per layer) and channel-wise quantizers for weights (one quantizer for each conv filter). The number of extra parameters introduced by the quantizers is negligible compared to the large volume of network weights.

### 3.3 Training Algorithm

To train the LQ-Nets, we use floating-point network weights which are quantized before convolution and optimized with error back-propagation (BP) and gradient descent. After training, they can be discarded and their binary codes and quantizer bases are kept. We now present how we optimize the quantizers.

**Quantizer optimization:** A simple and straightforward way to optimize our quantizers is through BP similar to weight optimization. Here we present an algorithm based on quantization error minimization which optimizes our quantizers in the forward passes during training. This algorithm leads to much better performance as we will show later in the experiments.

Let $\mathbf{x} = [x_1, ..., x_N]^{\mathrm{T}} \in \mathbb{R}^N$ be the full-precision data (weights or activations) and $K$ be the specified bit number for quantization. Our goal is to find an optimal quantizer basis $\mathbf{v} \in \mathbb{R}^K$ as well as an encoding $B = [\mathbf{b}_1, ..., \mathbf{b}_N] \in \{-1, 1\}^{K \times N}$ that minimize the quantization error:

$$\mathbf{v}^*, B^* = \arg\min_{\mathbf{v}, B} \left\| B^{\mathrm{T}} \mathbf{v} - \mathbf{x} \right\|_2^2, \quad s.t.\ B \in \{-1, 1\}^{K \times N}. \qquad (7)$$

Eq. (7) is complex and to provably solve for the optimal solution via brute-force search is exponential in the size of $B$. For efficiency purposes, we alternately solve for $\mathbf{v}$ and $B$ in a block coordinate descent fashion:

- *Fix $\mathbf{v}$ and update $B$.* Given $\mathbf{v}$, the optimal encoding $B^*$ can be simply found by looking up the quantization intervals $t_1, ..., t_{2^K+1}$.

---

**Algorithm 1** Training and testing the quantizers in LQ-Nets
---
1: **Parameters:** $\mathbf{v}$ - the current basis vector of the quantizer
2: **Input:** $\mathbf{x} = [x_1, x_2, ..., x_N]$ - the full-precision data (weights or activations)
3: **Output:** $B = [\mathbf{b}_1, ..., \mathbf{b}_N]$ - the binary encodings for the input data
4: **Procedure:**
5: **if** in the training stage **then** {//quantizer optimization with QEM in forward pass}
6:     Set $\mathbf{v}^{(0)} = \mathbf{v}$
7:     **for** $t = 1 \to T$ **do**
8:         Compute $B^{(t)}$ with $\mathbf{v}^{(t-1)}$ per Eq. (5)
9:         Compute $\mathbf{v}^{(t)}$ with $B^{(t)}$ per Eq. (8)
10:    **end for**
11:    Output $B = B^{(T)}$; Update current $\mathbf{v}$ with $\mathbf{v}^{(T)}$ via moving average
12: **else** {//simple quantization operation in the test stage}
13:    Compute $B$ with $\mathbf{v}$ per Eq. (5)
14: **end if**
---

– *Fix B and update* $\mathbf{v}$. Given $B$, Eq. (7) reduces to a linear regression problem with a closed form solution as

$$\mathbf{v}^* = (BB^{\mathrm{T}})^{-1}B\mathbf{x}. \quad (8)$$

We iterate the alternation $T$ times. For brevity, we will refer to the above procedure as the QEM (Quantization Error Minimization) algorithm.

**Network training:** We use the standard mini-batch based approach to train the LQ-Nets, and our quantizer learning is conducted in the forward passes with the QEM algorithm. Since, for activation quantization, only part of the input data is visible in one iteration due to mini-batch sampling, we apply moving average for the optimized quantizer parameters (i.e., basis vectors). We also apply the moving average strategy for the weight quantizers to gain more stability. The operations in our quantizers are summarized in Algorithm 1.

In a backward pass, direct error back-propagation would be problematic as the gradient of the quantization function is 0 at almost everywhere. To tackle this issue, we use the Straight-Through Estimator (STE) proposed in [2] to compute the gradients. Specifically, for activations we set the gradient of the quantization function to 1 for values between $q_1$ and $q_{2^K}$ defined in Eq. (5) and 0 elsewhere; for weights, the gradient is set to 1 everywhere [3]. The QEM algorithm is unrelated to the backward pass so the quantizers will remain unchanged (unless BP is used to train them instead).

## 4    Experiments

In this section, we evaluate the proposed method on two image classification datasets: CIFAR-10 [25] and ImageNet (ILSVRC12) [40]. The CIFAR-10 dataset consists of 60,000 color images of size $32 \times 32$ belonging to 10 classes (6,000 images per class). There are 50,000 training and 10,000 test images. ImageNet

ILSVRC12 contains about 1.2 million training and 50K validation images of 1,000 object categories.

Although our method is designed to quantize both weights and activations to facilitate fast inference through bitwise operations, we also conduct experiments of weight-only quantization and compare with the prior art.

### 4.1 Implementation Details

Our LQ-Nets are implemented with TensorFlow [1] and trained with the aid of the Tensorpack library [47].[4] We present our implementation details as follows.

**Quantizer implementation:** We apply layer re-ordering to the networks similar to [39, 3]: the typical Conv→BN→ReLU operations is re-organized as BN→ReLU (→Quant.)→Conv. Following previous methods [21, 39, 53, 54, 3], we quantize all the convolution and fully-connected layers but the first and last layers, for which the speedup benefited by bitwise operations is low due to their small channel number or filter size [39, 30].

**Network structures:** We conduct experiments on AlexNet [26], ResNet [15], DenseNet [20], two variants of the VGG-Net [41] "VGG-Small" and "VGG-Variant" from [3], and a variant of GoogLeNet [42] also from [3]. VGG-Small is a simplified VGG-Net similar to that of [21, 39] but with only one fully-connected layer. VGG-Variant is a smaller version of the model-A in [14]. The GoogLeNet structure in [3] contains some modifications of the original GoogLeNet (e.g., more filters in some $1 \times 1$ conv layers) and we denote it as "GoogleNet-Variant" in this paper. Detailed structures of these network variants can be found in [3]'s publicly-available implementation.[5] For ResNet-50, the parameter-free type-A shortcut [15] is adopted in this paper.

**Initialization:** In all the experiments, our LQ-Nets are trained from scratch (random initialization) without leveraging any pre-trained model. Our quantizers are initialized with uniform quantization (we also tried random initialization and initializing them via pre-computing the quantization levels using [3], however no noticeable difference on the results was observed).

**Hyper-parameters and other setup:** To train on various network architectures, we mostly follow the hyper-parameter settings (learning rate, batch size, training epoch, weight decay, etc.) of their original papers [15, 16, 20]. For fair comparisons with the method of [3], we use the hyper-parameters described in [3] to train all the networks with 1-bit weights and 2-bit activations. The iteration number $T$ in our QEM algorithm is fixed as 1 (no significant benefit was observed with larger values; see Section 4.2). The moving average factor for quantizer learning is fixed as 0.9. Details of all our hyper-parameter settings can be found in the *supplementary material* as well as our released source code.

---

[4] Our source code is available at https://github.com/Microsoft/LQ-Nets/
[5] https://github.com/zhaoweicai/hwgq (accessed July 10, 2018)

**Table 1:** Optimization method comparison on the ResNet-20 model.

| Bit-width (W/A) | Optim. method | Accuracy (%) |
|---|---|---|
| 2/32 | BP | 90.0 |
| 2/32 | QEM | **91.8** |
| 2/2 | BP | 88.2 |
| 2/2 | QEM | **90.2** |

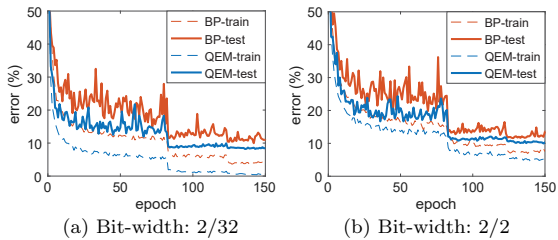

(a) Bit-width: 2/32  (b) Bit-width: 2/2

**Fig. 3:** Error curves with the two optimization methods

In the remaining text, we used "W/A" to denote the number of bits used for weights/activations. A bit-width of 32 indicates using 32-bit floating-point values without quantization (thus "$w/32$" with $w < 32$ indicates weight-only quantization and "32/32" are "full-precision" (FP) models). For the experiments on CIFAR-10, we run our method 5 times and report the mean accuracy.

### 4.2   Performance Analysis

**Effectiveness of the QEM algorithm:** Our quantizer can be trained by either the proposed QEM algorithm or a naive BP procedure. In this experiment, we evaluate the effectiveness of the QEM algorithm and compare it against BP. Table 1 shows the performance of the quantized ResNet-20 models on CIFAR-10 test set, and Fig. 3 presents the corresponding training and testing curves. The quantized network trained using QEM is clearly better than BP for weight-only quantization as well as weight-and-activation quantization. In all the following experiments, we use the QEM algorithm to optimize our quantizers.

Table 2 shows the accuracy of quantized ResNet-20 models with different QEM solver iteration $T$. As can be seen, using $T = 2, 3$ or 4 did not show significant benefit compared to $T = 1$. Note that each time the solver starts from the result of the last training iteration (see Line 6 in Algorithm 1) which is a good starting point especially when the gradients become small after a few epochs. The good performance with $T = 1$ suggests that the iterations of the alternately-directional optimization can be effectively substituted by the training iterations. In this paper, we use $T = 1$ in all the experiments.

**Table 2:** Accuracy w.r.t. QEM iteration number $T$

| Bit-width (W/A) | Accuracy (%; "mean±std" of 5 runs) | | | |
|---|---|---|---|---|
| | $T=1$ | $T=2$ | $T=3$ | $T=4$ |
| 1/2 | 88.37±0.26 | 88.38±0.11 | 88.45±0.12 | 88.51±0.20 |
| 2/2 | 90.16±0.08 | 90.35±0.12 | 90.33±0.20 | 90.25±0.14 |
| 3/3 | 91.58±0.16 | 91.55±0.23 | 91.62±0.15 | 91.44±0.14 |

**Effectiveness of the learnable quantizers:** The key idea of our method is to apply flexible quantizers and optimize them jointly with the network. Table 3 compares the results of our method and two previous methods: DoReFa-Net [53] and HWGQ [3], the former of which is based on fixed uniform quantizers and latter pre-computes the quantizer by fitting a half-wave Gaussian distribution. It can be seen that using 1-bit weights and 2-bit activations, the ResNet-18 model with our learnable quantizers outperformed HWGQ under the same setting and also outperformed DoReFa-Net with 4-bit activations on ImageNet. More result comparisons on various network structures can be found in Section 4.3.

**Table 3:** Comparison of different quantization methods (ResNet-18 on ImageNet)

| Method | Bit-width (W/A) | Accuracy (%) Top-1 | Top-5 |
|---|---|---|---|
| Uniform–DoReFa-Net [53] | 1/4 | 59.2 | - |
| Half-wave Gaussian–HWGQ [3] | 1/2 | 59.6 | 82.2 |
| *Ours* | 1/2 | **62.6** | **84.3** |

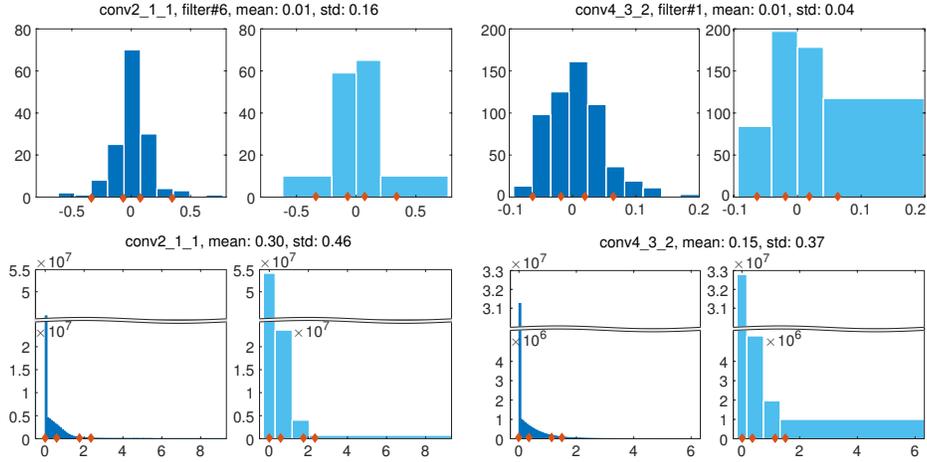

**Fig. 4:** Statistics of the weights (top row) and activations (bottom row) before (i.e., the floating-point values) and after quantization. A ResNet-20 model trained on CIFAR-10 with "2/2" quantization is used. The orange diamonds indicate the four quantization levels of our learned quantizers. Note that in the left figures for the floating-point values, the histogram bins are of equal step size, whereas in the right figures each of the four bins contains all the values quantized to its corresponding quantization levels.

Figure 4 presents the weight and activation statistics in two layers of a trained ResNet-20 model before (i.e., the floating-point values) and after quantization using our method. The network is quantized with "2/2" bits and the floating-point weights are obtained from the last iteration of training (these values can be discarded after training and only quantized values are used in the inference time). The floating-point activations are obtained using all the test images of CIFAR-10. It can be seen that our learned quantizers are not uniform ones and they differ at different layers. Statistical results with more bits can be found in the *supplementary material*.

**Performance w.r.t. Bit-width:** We now study the impact of bit-width on the performance of our LQ-Nets. Table 4 shows the results of three network structures: ResNet-20, VGG-Small and ResNet-18.

On the CIFAR-10 dataset, high accuracy can be achieved by our low-precision networks. The accuracy from "3/32" quantization has roughly reached our full-precision result for both ResNet-20 and VGG-Small. The accuracy decreases gracefully with lower bits for weights, and the absolution drops are low even with 1-bit weights: 2.0% for ResNet-20 and 0.3% for VGG-Small. The accuracy drops are more appreciable when quantizing both weights and activations, though the largest absolute drop is only 3.7% for the "1/2" ResNet-20 model. Very minor accuracy drops (maximum 0.4%) are observed for VGG-Small which has many more parameters than ResNet-20.

Table 4: Impact of bit-width on our LQ-Nets

| ResNet-20 (CIFAR-10) | | VGG-Small (CIFAR-10) | | ResNet-18 (ImageNet) | |
|---|---|---|---|---|---|
| Bit-width (W/A) | Acc. (%) | Bit-width (W/A) | Acc. (%) | Bit-width (W/A) | Acc. (%) |
| 32/32 | 92.1 | 32/32 | 93.8 | 32/32 | 70.3 |
| 1/32 | 90.1 | 1/32 | 93.5 | 2/32 | 68.0 |
| 2/32 | 91.8 | 2/32 | 93.8 | 3/32 | 69.3 |
| 3/32 | 92.0 | 3/32 | 93.8 | 4/32 | 70.0 |
| 1/2 | 88.4 | 1/2 | 93.4 | 1/2 | 62.6 |
| 2/2 | 90.2 | 2/2 | 93.5 | 2/2 | 64.9 |
| 2/3 | 91.1 | 2/3 | 93.8 | 3/3 | 68.2 |
| 3/3 | 91.6 | 3/3 | 93.8 | 4/4 | 69.3 |

On the ImageNet dataset which is more challenging, the accuracy drops of the ResNet-18 model after quantization are relatively larger especially with very low precision: the largest absolute drop is 7.7% (70.3%→62.6%) with bit-widths of "1/2". Nevertheless, our learnable quantizer is particularly beneficial when using 2 or more bits due to its high flexibility. The accuracy of the quantized ResNet-18 quickly increases with 2 or more bits as shown in Table 4. The accuracy gap is almost closed with "4/32" bits (0.3% absolute difference only), and the accuracy drop with the "4/4" case is as low as 1%.

### 4.3 Comparison with Previous Methods

In this section, we compare the performance of our quantization method with existing methods including TWN [29], TTQ [54], BNN [21], BWN [39], XNOR-Net [39], DoReFa-Net [53], HWGQ [3] and ABC-Net [33], with various network architectures tested on CIFAR-10 and ImageNet classification tasks.

**Comparison on CIFAR-10:** Table 5 presents the results of the VGG-Small model quantized using different methods. All these methods quantize (or binarize) both weights and activations to achieve extremely low precision. With 1-bit weights and 2-bit activations, the accuracy using our method is significantly better than the state-of-the-art method HWGQ (93.4% vs. 92.5%).

Table 5: Comparison of quantized VGG-Small networks on CIFAR-10

| Model | Methods | Bit-width (W/A) | Acc. (%) |
|---|---|---|---|
| VGG-Small | FP-$_{\text{HWGQ}}$ [3] | 32/32 | 93.2 |
| | FP-ours | 32/32 | 93.8 |
| | BNN [21] | 1/1 | 89.9 |
| | XNOR-Net [39] | 1/1 | 89.8 |
| | HWGQ [3] | 1/2 | 92.5 |
| | *Ours* | 1/2 | **93.4** |

**Table 6:** Comparison with state-of-the-art quantization methods on ImageNet. "FP" denotes "Full Precision"; the "W/A" values are the bit-widths of weights/activations.

| Methods | Bit-width (W/A) | Accuracy(%) Top-1 | Top-5 |
|---|---|---|---|
| ResNet-18 | | | |
| FP* [15] | 32/32 | 69.6 | 89.2 |
| FP-ours | 32/32 | 70.3 | 89.5 |
| BWN [39] | 1/32 | 60.8 | 83.0 |
| TWN [29] | 2/32 | 61.8 | 84.2 |
| TWN[†] [29] | 2/32 | 65.3 | 86.2 |
| TTQ[†] [54] | 2/32 | 66.6 | 87.2 |
| Ours | 2/32 | **68.0** | **88.0** |
| Ours | 3/32 | **69.3** | **88.8** |
| Ours | 4/32 | **70.0** | **89.1** |
| XNOR-Net [39] | 1/1 | 51.2 | 73.2 |
| DoReFa-Net[‡] [53] | 1/2 | 53.4 | - |
| DoReFa-Net[‡] [53] | 1/4 | 59.2 | - |
| HWGQ [3] | 1/2 | 59.6 | 82.2 |
| ABC-Net [33] | 3/3 | 61.0 | 83.2 |
| ABC-Net [33] | 5/5 | 65.0 | 85.9 |
| Ours | 1/2 | **62.6** | **84.3** |
| Ours | 2/2 | **64.9** | **85.9** |
| Ours | 3/3 | **68.2** | **87.9** |
| Ours | 4/4 | **69.3** | **88.8** |
| ResNet-34 | | | |
| FP* [15] | 32/32 | 73.3 | 91.3 |
| FP-ours | 32/32 | 73.8 | 91.4 |
| HWGQ [3] | 1/2 | 64.3 | 85.7 |
| ABC-Net [33] | 3/3 | 66.7 | 87.4 |
| ABC-Net [33] | 5/5 | 68.4 | 88.2 |
| Ours | 1/2 | **66.6** | **86.9** |
| Ours | 2/2 | **69.8** | **89.1** |
| Ours | 3/3 | **71.9** | **90.2** |
| ResNet-50 | | | |
| FP* [15] | 32/32 | 76.0 | 93.0 |
| FP-ours | 32/32 | 76.4 | 93.2 |
| Ours | 2/32 | **75.1** | **92.3** |
| Ours | 4/32 | **76.4** | **93.1** |
| HWGQ [3] | 1/2 | 64.6 | 85.9 |
| ABC-Net [33] | 5/5 | 70.1 | 89.7 |
| Ours | 1/2 | **68.7** | **88.4** |
| Ours | 2/2 | **71.5** | **90.3** |
| Ours | 3/3 | **74.2** | **91.6** |
| Ours | 4/4 | **75.1** | **92.4** |

| Methods | Bit-width (W/A) | Accuracy(%) Top-1 | Top-5 |
|---|---|---|---|
| AlexNet | | | |
| FP [26] | 32/32 | 57.1 | 80.2 |
| FP-ours | 32/32 | 61.8 | 83.5 |
| BWN [39] | 1/32 | 56.8 | 79.4 |
| DoReFa-Net[§] [53] | 1/32 | 53.9 | 76.3 |
| TWN[§] [29] | 2/32 | 54.5 | 76.8 |
| TTQ [54] | 2/32 | 57.5 | 79.7 |
| Ours | 2/32 | **60.5** | **82.7** |
| BNN[§] [21] | 1/1 | 41.8 | 67.1 |
| XNOR-Net [39] | 1/1 | 44.2 | 69.2 |
| DoReFa-Net [53] | 1/1 | 43.6 | - |
| DoReFa-Net [53] | 1/2 | 49.8 | - |
| DoReFa-Net [53] | 1/4 | 53.0 | - |
| HWGQ [3] | 1/2 | 52.7 | 76.3 |
| Ours | 1/2 | **55.7** | **78.8** |
| Ours | 2/2 | **57.4** | **80.1** |
| DenseNet-121 | | | |
| FP [20] | 32/32 | 75.0 | 92.3 |
| FP-ours | 32/32 | 75.3 | 92.5 |
| DoReFa-Net* [53] | 2/2 | 67.7 | 88.4 |
| Ours | 2/2 | **69.6** | **89.1** |
| VGG-Variant | | | |
| FP-HWGQ [3] | 32/32 | 69.8 | 89.3 |
| FP-ours | 32/32 | 72.0 | 90.5 |
| HWGQ [3] | 1/2 | 64.1 | 85.6 |
| Ours | 1/2 | **67.1** | **87.6** |
| Ours | 2/2 | **68.8** | **88.6** |
| GoogLeNet-Variant | | | |
| FP-HWGQ [3] | 32/32 | 71.4 | 90.5 |
| FP-ours | 32/32 | 72.9 | 91.3 |
| HWGQ [3] | 1/2 | 63.0 | 84.9 |
| Ours | 1/2 | **65.6** | **86.4** |
| Ours | 2/2 | **68.2** | **88.1** |

---

* Results of the ResNet models trained in Torch: https://github.com/facebook/fb.resnet.torch (accessed July 10, 2018)
[†] Results of ResNet-18B [29] where the filter number in each block is 1.5× of ResNet-18
[‡] Results quoted from the method's official webpage https://github.com/tensorpack/tensorpack/tree/master/examples/DoReFa-Net (accessed July 10, 2018)
⋆ Results trained by us with the authors' code (same URL as above)
[§] Results quoted from [54] for TWN and DoReFa-Net, and [22] for BNN.

**Comparison on ImageNet:** The results on ImageNet validation set are presented in Table 6. For weight-only quantization, our LQ-Nets outperformed BWN, TWN, TTQ and DoReFa-Net by large margins.

As for quantizing both weights and activations, our results are significantly better than DoReFa-Net and HWGQ when using very low bit-widths (1 bit for weights and few for activations). Our method is even more advantageous when using larger bit-widths. Table 6 shows that with more bits (2, 3, or 4), the accuracy can be dramatically improved by our method. For example, with "4/4" bits, the top-1 accuracy of ResNet-50 is boosted from 68.7% (with "1/2" bits) to 75.1%. The absolute accuracy increase is as high as 6.4%, and the gap to its FP counterpart is reduced to 1.3%. According to Table 6, the accuracy of our LQ-Nets comprehensively surpassed the other competing methods under the same bit-width settings.

### 4.4   Training time

Compared to training floating-point networks, our extra cost lies in quantizer optimization. In the QEM algorithm, the cost of solving $B$ is negligible. For $N$ input scalars, the time complexity of solving $\mathbf{v}$ of length $K$ is $O(K^2 N)$, which is a relatively small compared to the conv operations in theory.[6] Table 7 shows the total training time comparison based on our current unoptimized implementation. The network is ResNet-18 and no bitwise operation is used in all cases. Our training time increases gracefully with larger bit-widths.

**Table 7:** Train. time

| Bit-width (W/A) | Training time |
|---|---|
| 32/32 | 1.0× |
| 2/32 | 1.4× |
| 3/32 | 1.7× |
| 1/2 | 2.1× |
| 2/2 | 2.3× |
| 3/3 | 3.7× |

## 5   Conclusions

We have presented a novel DNN quantization method that led to state-of-the-art accuracy for various network structures. The key idea is to apply learnable quantizers which can be jointly trained with the network parameters to gain more flexibility. Our quantizers can be applied to both weights and activations, and they are made compatible with bitwise operations facilitating fast inference. In future, we plan to deploy our LQ-Nets on some resource-constrained devices such as mobile phones and test their performance.

**Acknowledgment**   This work is partially supported by the National Natural Science Foundation of China under Grant 61629301.

---

[6] To solve for $\mathbf{v}$ in Eq. (8), we need $O(K^2 N)$ for matrix multiplication $BB^{\mathrm{T}}$, $O(K^3)$ for matrix inverse, and $O(KN)$ for matrix-vector multiplications. Note $K \ll N$. Let the input and output activation map sizes be $(H, W, C_{in})$ and $(H, W, C_{out})$. The input activation number is $N_a = C_{in} H W$. The time complexity of an $S \times S$ conv operation with stride 1 is $O(S^2 C_{out} C_{in} H W) = O(S^2 C_{out} N_a)$, whereas that of the quantizer optimization is $O(K_a^2 N_a)$ for activations and $O(K_w^2 N_w)$ for weights.

# LQ-Nets: Learned Quantization for Highly Accurate and Compact Deep Neural Networks (*Supplementary Material*)

## 1 Statistics of Weights and Activations

In the main paper we presented the statistics of weights and activations in the ResNet-20 model quantized with "2/2" bits. Here we show the cases with "3/3" bit-widths in Fig. I.

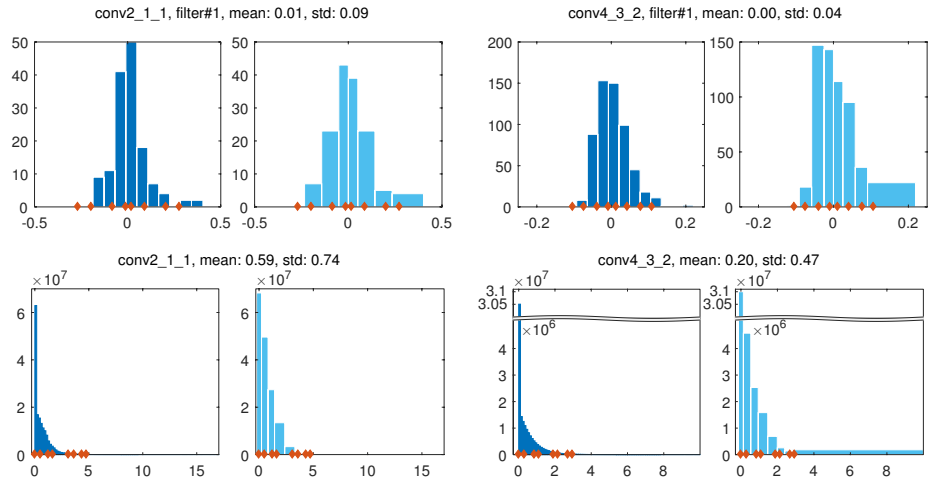

**Fig. I:** Statistics of the weights (top row) and activations (bottom row) before (i.e., the floating-point values) and after quantization. The ResNet-20 model with "3/3" quantization is used. The orange diamonds indicate the eight quantization levels of our learned quantizers. Note that in the left figures for the floating-point values the histogram bins are of equal step size, whereas in the right figures each of the four bins contains all the values quantized to its corresponding quantization levels.

## 2 Detailed Hyper-Parameter and Other Setups

We presented here the detailed hyper-parameters and other training setups that are omitted in the main paper due to space limitation.

### 2.1 CIFAR-10 Experiments

**Data augmentation:** Following [4, 2], in the training stage we pad 4 pixels on each side of the original 32×32 images, and randomly crop a 32×32 sample or its horizontal flip. The original images are used at test time.

**Hyper-parameters:** For all the experiments on CIFAR-10, we train the models for up to 200 epochs and use a momentum of 0.9. For the ResNet-20 model, the learning rate starts at 0.1 and is divided by 10 at 82 and 123 epochs. Weight decay of $1e-4$ and batch size of 128 are adopted following the original paper. For VGG-Small, the learning rate starts at 0.02 and is divided by 10 at 80 and 160 epochs. Following [1], we set weight decay to $5e-4$ and batch size to 100.

### 2.2 ImageNet Experiments

**Data augmentation:** Our data augmentation strategy mostly follows [1]. During training, we first resize the shorter side of the images to 256, and then randomly sample 224×224 (227×227 for AlexNet) image crops with horizontal flipping applied at random. At test time, a single, centered crop of size 224×224 (227×227 for AlexNet) is used for each image. When training networks with bit-widths larger than "1/2", we follow the augmentation strategy of the ResNet Torch implementation[1]. Specifically, we use the scale and aspect ratio augmentation from [5] and color augmentation proposed in [3].

**Hyper-parameters:** For all the experiments on ImageNet, following [2] we train the models for up to 120 epochs with a momentum of 0.9. For all the experiments with bit-widths larger than "1/2", the batch size is 256 and the weight decay is $1e-4$. The learning rate starts at 0.1 and is divided by 10 at 30, 60, 85, 95, 105 epochs.

When comparing against HWGQ [1] on ResNet, AlexNet, VGG-Variant and GoogLeNet with bit-widths of "1/2", we use the same hyper-parameters in HWGQ's implementation. Specifically, the learning rate starts at 0.1 for ResNet and GoogLeNet, 0.01 for VGG-Variant, and 0.02 for AlexNet, respectively. Polynomial learning rate annealing with power of 1 is adopted instead of the multi-step annealing. The total training epoch is set to 64 for all experiments. The batch size is 128 for VGG-Variant and 256 for others. The weight decay is $5e-4$ for AlexNet and VGG-Variant, and $5e-5$ for ResNet and GoogLeNet.

---

[1] https://github.com/facebook/fb.resnet.torch (accessed July 10, 2018)